\documentclass{article}

\usepackage[dblblindworkshop, final]{neurips_2025}
\usepackage{amsmath}
\usepackage{natbib}

\usepackage[utf8]{inputenc} 
\usepackage[T1]{fontenc}    
\usepackage{hyperref}       
\usepackage{url}            
\usepackage{booktabs}       
\usepackage{amsfonts}       
\usepackage{nicefrac}       
\usepackage{microtype}      
\usepackage{xcolor}         

\usepackage{multirow}
\usepackage{graphicx}
\graphicspath{ {./images/} }

\usepackage{wrapfig,lipsum}
\usepackage{float}
\usepackage{multirow}
\usepackage{arydshln} 

\definecolor{sapblue}{HTML}{0056D0}
\definecolor{sapgreen}{HTML}{00C9AF}
\definecolor{sappurple}{HTML}{8F57C8}
\usepackage{caption}
\usepackage{listings}

\lstdefinestyle{json}{
    basicstyle=\footnotesize\ttfamily,
    numbers=left,
    numberstyle=\tiny,
    stepnumber=5,
    numbersep=8pt,
    showstringspaces=false,
    breaklines=true,
    frame=lines,
    backgroundcolor=\color{white!10},
    literate=
     *{0}{{{\color{sapgreen}0}}}{1}
      {1}{{{\color{sapgreen}1}}}{1}
      {2}{{{\color{sapgreen}2}}}{1}
      {3}{{{\color{sapgreen}3}}}{1}
      {4}{{{\color{sapgreen}4}}}{1}
      {5}{{{\color{sapgreen}5}}}{1}
      {6}{{{\color{sapgreen}6}}}{1}
      {7}{{{\color{sapgreen}7}}}{1}
      {8}{{{\color{sapgreen}8}}}{1}
      {9}{{{\color{sapgreen}9}}}{1}
      {:}{{{\color{sapblue}{:}}}}{1}
      {,}{{{\color{sapblue}{,}}}}{1}
      {\{}{{{\color{sappurple}{\{}}}}{1}
      {\}}{{{\color{sappurple}{\}}}}}{1}
      {[}{{{\color{sappurple}{[}}}}{1}
      {]}{{{\color{sappurple}{]}}}}{1},
    string=[s]{"}{"},
    stringstyle=\color{sapgreen!60!black},
    comment=[l]{//},
    commentstyle=\color{gray},
}

\title{SALT-KG: A Benchmark for Semantics-Aware Learning on Enterprise Tables}

\author{%
  Isaiah Onando Mulang' \\
  SAP SE\\
  \texttt{\small mulang.onando@sap.com} \\
  \And
  Felix Sasaki \\
  SAP SE \\
  \texttt{\small felix.sasaki@sap.com} \\
  \And
  Tassilo Klein \\
  SAP SE \\
  \texttt{\small tassilo.klein@sap.com} \\
  \And
  Jonas Kolk \\
  SAP SE \\
  \texttt{\small jonas.kolk@sap.com} \\
  \And
  Nikolay Grechanov \\
  SAP SE \\
  \texttt{\small nikolay.grechanov@sap.com} \\
   \And
  Johannes Hoffart \\
  SAP SE \\
  \texttt{\small johannes.hoffart01@sap.com} \\
}

\begin{document}
\maketitle
\begin{abstract}
Building upon the \textsc{SALT} benchmark for relational prediction~\citep{klein2024salt}, we introduce \textsc{SALT-KG}, a benchmark for semantics-aware learning on enterprise tables. 
\textsc{SALT-KG} extends \textsc{SALT} by linking its multi-table transactional data with a structured \textbf{\underline{O}}perational \textbf{\underline{B}}usiness Knowledge represented in a Metadata \textbf{\underline{K}}nowledge \textbf{\underline{G}}raph (\textbf{OBKG}) that captures field-level descriptions, relational dependencies, and business object-types. 
This extension enables evaluation of models that jointly reason over tabular evidence and contextual semantics—an increasingly critical capability for foundation models on structured data. 
Empirical analysis reveals that while metadata-derived features yield modest improvements in classical prediction metrics, these metadata features consistently highlight gaps in models’ ability to leverage semantics in relational context. 
By reframing tabular prediction as semantics-conditioned reasoning, \textsc{SALT-KG} establishes a benchmark to advance tabular FMs grounded in declarative knowledge, providing the first empirical step toward semantically linked tables in structured data at enterprise scale.\footnote{\tiny The data set, benchmark splits, and scripts to reproduce all results are publicly available at: https://github.com/SAP-samples/salt-kg.
}
\end{abstract}

\vspace{-5pt}
\section{Introduction}
\vspace{-5pt}
We introduce \textsc{SALT-KG}, a benchmark for \emph{semantics-aware learning on enterprise tables}. The \textsc{SALT} dataset~\citep{klein2024salt} established a standardized setting for learning multi-table enterprise data through attribute autocompletion tasks, but was limited to relational structure, focusing on how models infer missing attributes from transactional evidence.  \textsc{SALT-KG} extends this foundation by linking the same relational schema to a structured OBKG that captures field-level semantics, declarative business knowledge, and hierarchical object types. 
Declarative metadata describes what entities and attributes \emph{mean}, beyond how they are \emph{joined}, enabling models to condition tabular prediction on contextual semantics instead of mere statistical correlations. \\
Despite remarkable progress in language, vision, and multimodal learning, tabular data—particularly multi-table, relational datasets typical of enterprise environments—remain among the most challenging modalities for machine learning~\citep{grinsztajn2022tree, bodensohn2024llms}. 
Recent advances in tabular foundation models, such as \textsc{CARTE}~\citep{kim2024carte}, \textsc{TabPFN}~\citep{hollmann2025tabpfn,hollmann2023tabpfn}, \textsc{TabICL}~\citep{qu2025tabicl}, \textsc{PORTAL}~\citep{spinaci2024portal} and \textsc{ContextTab}~\citep{spinaci2025contexttab}, have improved data efficiency and in-context reasoning. 
However, these models are typically trained and evaluated on benchmarks that represent relational structure but lack explicit semantic grounding or declarative context. 
Conversely, enterprise data encodes rich dependencies between attributes—such as \textit{Shipping Point}, \textit{Plant}, and \textit{Payment Terms}—yet the semantics of these relationships are implicit, buried in metadata descriptions or business logic. 
As a result, even strong tree-based or neural predictors rely on the memorization of local correlations rather than reasoning. \\
Existing benchmarks such as \textsc{RelBench}~\citep{relbench}, \textsc{TALENT}~\citep{liu2024talenttabularanalyticslearning}, \textsc{SALT}~\citep{klein2024salt}, and \textsc{TabArena}~\citep{erickson2025tabarenalivingbenchmarkmachine} have advanced evaluation for structured prediction on relational data. 
Yet, they do not expose the metadata layer where domain meaning and declarative knowledge reside. 
In parallel, knowledge graph (KG) and data integration communities have explored connecting tables to semantic graphs through systems such as \textsc{JenTab}~\citep{ermilov2023jentab}, \textsc{CoreColumnMatch}~\citep{li2024corecolumn}, and LLM-based table-to-KG alignment methods~\citep{zheng2024scalable}. 
These advances have yet to be systematically incorporated into tabular learning benchmarks, leaving a gap between relational representation learning and semantics-aware reasoning. \textsc{SALT-KG} bridges this gap by enriching enterprise relational data with an explicit semantic layer that links tables, fields, and business object types through declarative relationships in a coherent OBKG. 
This integration enables controlled evaluation of models that combine relational reasoning with declarative schema semantics for contextual understanding beyond feature-level learning.
Empirical results show that metadata features yield only modest gains in classical prediction metrics but reveal consistent differences in the models' ability to exploit semantic alignment and contextual relationships. 
This work operationalizes the declarative layer of the broader \emph{Foundation Models for Semantically Linked Tables (FMSLT)} framework~\citep{klein2025foundationmodelstabulardata}, which envisions integrating declarative, procedural, and operational knowledge; \textsc{SALT-KG} focuses solely on the declarative dimension—capturing what entities and attributes \emph{mean} through structured metadata rather than how they \emph{behave} in processes or systems.

\noindent{\textbf{Related Work:}}
\label{related-work}
The prediction of tabular data has traditionally been dominated by tree-based ensemble methods such as XGBoost~\citep{chenXGBoost}, LightGBM~\citep{LightGBMKe}, and CatBoost~\citep{CATBoostProkhorenkova}. 
Recently, this landscape has shifted with the emergence of deep learning approaches tailored to relational databases, including \textsc{CARTE}~\citep{kim2024carte}, AutoGluon~\citep{erickson2020autogluontabularrobustaccurateautoml}, and GraphSAGE~\citep{hamilton2017inductive}, evaluated on benchmarks such as RelBench~\citep{relbench} and TabArena~\citep{erickson2025tabarenalivingbenchmarkmachine}. 
The rise of tabular foundation models—including transformer-based in-context learners such as TabPFN~\citep{hollmann2025tabpfn,hollmann2023tabpfn}, TabICL~\citep{qu2025tabicl}, and ContextTab~\citep{spinaci2025contexttab}—marks a significant shift toward generalizable tabular learning. 
In parallel, knowledge graphs (KGs) have gained prominence for their ability to contextualize data for deep learning~\citep{mulang2020impact}. 
Based on this, systems such as JenTab~\citep{ermilov2023jentab} have taken initial steps toward bridging the gap between raw tables and semantic graphs. 
Nevertheless, more progress is needed to ground tabular foundation models in operational business semantics~\citep{klein2025foundationmodelstabulardata}, enabling models that jointly reason over relational evidence and contextual knowledge. 
SALT-KG is designed to catalyze this research direction by providing a benchmark to evaluate how contextual knowledge of KG can improve tabular learning and representation.

\vspace{-5pt}
\section{Dataset Design}
\vspace{-5pt}
\label{salt-kg}

\noindent{\textbf{Background:}} 
SALT-KG extends \textsc{SALT}~\citep{klein2024salt} by enriching multi-table enterprise data with a structured layer of KG-metadata. 
The dataset captures a representative \textit{sales-order creation process} from a real-world transactional system, where each order consists of a document header and multiple line items linked through foreign keys. 
This reflects enterprise data characteristics: multi-relational dependencies, temporal dynamics, categorical imbalance, and hierarchical structure. 

\noindent{\textbf{Task Definition:}}
The SALT-KG benchmark defines a suite of predictive tasks designed to assess \emph{semantics-aware tabular learning}. 
Aligned with \textsc{SALT} setup~\citep{klein2024salt}, the benchmark simulates missing-field autocompletion in transactional records, to infer key attributes given partial table information. 
Twenty-one features are provided as inputs, and eight variables serve as multiclass targets, including sales indicators such as \texttt{SalesOffice}, \texttt{SalesGroup}, \texttt{CustomerPaymentTerms}, \texttt{ShippingCondition}, \texttt{Plant}, \texttt{ShippingPoint}, and both header- and item-level \texttt{IncotermsClassification}. 

\begin{figure}[t]
\includegraphics[width=1.0\textwidth]{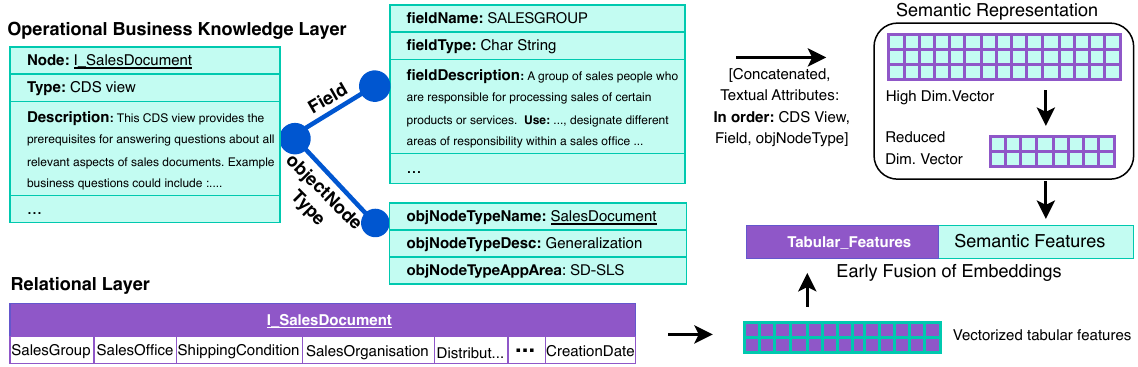}
\caption{Construction of semantically linked tables. 
A sample OBKG (blue)
aligned with its relational schema. 
Graph context is encoded and merged with tabular features.}

\label{fig:kg-semantics}
\end{figure}

The OBKG data infused with the tabular SALT is obtained from an underlying RDF-Based Enterprise Metadata KG. The KG models the ontological concepts for business data, and instances of these concepts are represented in a hierarchical network beginning with metadata about the underlying tables (e.g., table names, descriptions, column descriptions, type, length) and other metadata such as property domains and ranges that define RDF classes to which these objects belong. For example, in the table "\textsc{I\_SalesDocument}" will be represented as a node in the graph as an instance of the class for tables. The fields are independent one-hop nodes,  allowing for more expressivity at the table and field levels. Further metadata includes value help data and the business application area from where the table has been derived. The next layer in the network involves inter-table relationships, specifically those involving foreign keys and joins within the network. Core Data Services Views (known as \textsc{CDS Views}\footnote{\tiny https://help.sap.com/docs/SAP\_HANA\_PLATFORM/09b6623836854766b682356393c6c416/b4080c0883c24d2dae38a60d7fbf07c8.html})
are projections of other entities, like tables, to capture specific data needs. In the OBKG, these are data abstraction nodes with associated fields, labels, associations, data classes, reference fields, and other elements. Intrinsically, every underlying table has a matching view in the KG. To align OBKG with the underlying tabular data, metadata context is fetched using \textsc{SPARQL} queries in which KG nodes corresponding to each of the tables in SALT are subjects of the first triple patterns in the SPARQL basic graph pattern.  Triple patterns for one-hop triples that link to two major business objects, namely: The Fields and the Object Node Types are fetched. Fig. \ref{fig:kg-semantics} illustrates three types of interconnected information. First, the CDS Views provides data model abstractions (semantic definitions without actual tabular instances. Second, the fields augment the columns in the SALT table, e.g., \textsc{I\_SalesDocument} in our example has a column \textsc{AccountingDocExternalReference} and the OBKG triples provide descriptions or data type information for this column. column. All tabular columns have a corresponding field in the KG. Third, Object Node Types provide further semantic metadata through technical definitions, business object descriptions, and additional  information such as the application area of the business object.\\ 

\vspace{-5pt}

The overall dataset comprises exact tabular statistics in SALT~\citep{klein2024salt}. 
These tables are normalized through foreign-key joins and flattened to the item level for supervised learning.  Approximately 990 schema fields are explicitly mapped to corresponding tabular columns, along with their associated business object nodes and textual descriptions.  Additionally, 1{,}954 semantic object types enrich the metadata layer. Connectivity varies across schema elements: object nodes related to address entities exhibit higher out-degree, as address views connect to numerous other schema components across diverse contexts. These nodes reference a shared set of higher-level objects. Details of extracted semantic information available in appendix \ref{apdx:data-stats}. This semantic overlay transforms the relational dataset into a graph-linked benchmark, enabling contextual representation learning and semantics-aware evaluation across tabular and graph modalities.

\noindent{\textbf{Data Insights and Challenges:}}
SALT-KG inherits the same real-world properties of \textsc{SALT}~\citep{klein2024salt}: high cardinality, class imbalance, and temporal drift. 
Several categorical attributes, such as \texttt{SalesOffice} and \texttt{ShippingPoint}, are dominated by few frequent classes, whereas others (e.g., \texttt{Customer} or \texttt{ProductID}) contain tens of thousands distinct values. 
Temporal drift is evident across the three-year window, as organizational structures and categorical hierarchies evolve with time. 
Introducing the KG layer opens new opportunities to address these challenges. 
By linking tabular fields to their semantic object types and descriptive definitions, models can leverage graph-based regularization, hierarchical grouping, or transfer between semantically related columns. E.g. graph relations connecting address-related entities across schema components like \texttt{Customer}, \texttt{AddressLine}, \texttt{PostalLocation} can facilitate domain adaptation or infer meaningful join paths between related tables. 
This design encourages research on semantics-informed learning strategies that transcend feature engineering, emphasizing generalization through contextual and relational understanding.
\hspace{-5pt}
\begin{table}[htbp]
\tiny
\centering
\caption{\textbf{Effect of KG context on baseline models ($\Delta$MRR).}
\textcolor{sapblue}{Blue} values indicate improvement due to KG context; 
\textcolor{red}{red} values indicate degradation; black values denote no change.
}

\renewcommand{\arraystretch}{1.2}
\setlength{\tabcolsep}{3pt}

\resizebox{\textwidth}{!}{%
\begin{tabular}{@{}l *{9}{c}@{}} 
\toprule
\textbf{Method \textbackslash{} Target} &
\textbf{Plant} &
\textbf{ShipPt} &
\textbf{Item Inc.} &
\textbf{Hdr Inc.} &
\textbf{Sales Off.} &
\textbf{Sales Grp.} &
\textbf{PayTrm} &
\textbf{ShipCond} &
\textbf{Avg $\Delta$} \\
\midrule

Random Classifier & 
\textcolor{sapblue}{+0.03} &
\textcolor{sapblue}{+0.02} &
0.00 &
0.00 &
\textcolor{sapblue}{+0.01} &
\textcolor{red}{-0.01} &
\textcolor{red}{-0.02} &
\textcolor{sapblue}{+0.01} &
\textcolor{sapblue}{+0.01} \\

Majority Class Baseline &
\textcolor{sapblue}{+0.03} &
\textcolor{red}{-0.05} &
0.00 &
0.00 &
\textcolor{sapblue}{+0.01} &
\textcolor{red}{-0.01} &
0.00 &
0.00 &
\textcolor{sapblue}{+0.00} \\

\hdashline
XGBoost (~\citep{chenXGBoost}) &
0.00 &
0.00 &
\textcolor{sapblue}{+0.01} &
\textcolor{sapblue}{+0.01} &
\textcolor{sapblue}{+0.01} &
0.00 &
0.00 &
\textcolor{sapblue}{+0.01} &
\textcolor{sapblue}{+0.01} \\

LightGBM (~\citep{LightGBMKe}) &
\textcolor{sapblue}{+0.01} &
0.00 &
\textcolor{sapblue}{+0.03} &
\textcolor{sapblue}{+0.04} &
\textcolor{sapblue}{+0.01} &
\textcolor{sapblue}{+0.01} &
\textcolor{sapblue}{+0.08} &
\textcolor{red}{-0.31} &
\textcolor{sapblue}{+0.00} \\

CatBoost (~\citep{CATBoostProkhorenkova}) &
0.00 &
\textcolor{red}{-0.09} &
\textcolor{red}{-0.02} &
0.00 &
\textcolor{sapblue}{+0.01} &
0.00 &
0.00 &
0.00 &
\textcolor{red}{-0.01} \\

\hdashline
CARTE (~\citep{kim2024carte}) &
0.00 &
\textcolor{sapblue}{+0.01} &
\textcolor{sapblue}{+0.02} &
\textcolor{sapblue}{+0.02} &
\textcolor{sapblue}{+0.01} &
\textcolor{red}{-0.01} &
\textcolor{sapblue}{+0.03} &
\textcolor{sapblue}{+0.04} &
\textcolor{sapblue}{+0.02} \\

AutoGluon (~\citep{erickson2020autogluontabularrobustaccurateautoml}) &
0.00 &
0.00 &
\textcolor{sapblue}{+0.02} &
\textcolor{sapblue}{+0.03} &
\textcolor{sapblue}{+0.01} &
\textcolor{sapblue}{+0.02} &
\textcolor{red}{-0.04} &
0.00 &
\textcolor{sapblue}{+0.03} \\

GraphSAGE (~\citep{hamilton2017inductive}) &
0.00 &
\textcolor{sapblue}{+0.01} &
0.00 &
\textcolor{red}{-0.01} &
\textcolor{sapblue}{+0.01} &
\textcolor{red}{-0.02} &
\textcolor{sapblue}{+0.01} &
\textcolor{red}{-0.10} &
\textcolor{red}{-0.01} \\
\bottomrule
\end{tabular}
}
\label{tab-results}
\end{table}

\vspace{-5pt}
\section{Experiments and Results}
\vspace{-5pt}

\vspace{-5pt}
We evaluate \textsc{SALT-KG} using standard tabular baselines trained on the joined relational schema, similar to the setup in SALT ~\cite{klein2024salt}. 
Since OBKG provides schema rather than record level information, we independently encode the  information (concatenated in the order: CDS View, Fields, objNodeTypes), using  \textsc{text-embedding-3-large} model, which we chose as baseline method for simplicity of implementation.
The encoded descriptors are then projected into a compact latent space via PCA. We perform early fusion of the two representations, where tabular features are encoded into unified row vectors by retaining both the numerical and categorical features for tree based methods, while relying on intrinsic representations for the neural models. The resultant reduced-dimensional semantic representation is concatenated with tabular features for every row. This final representation is then used to train all eight benchmark models. 
Empirically, retaining between 16 and 64 components provides a stable trade-off for expressiveness and generalization, while higher dimensions may lead to noise amplification, weak signal utilization, and increased feature space complexity that hinders robust learning leading to degraded performance.

Deep learning baselines benefit modestly from this semantic context, while graph-based models, such as \textsc{GraphSAGE}~\citep{hamilton2017inductive}, show  inconsistency, highlighting the need for architectures that directly integrate declarative schema semantics. 
Across all baselines, incorporating metadata-derived features rarely changes overall ranking metrics but consistently alters the relative performance of different model families, as shown by the slight deviations in Tab.~\ref {tab-results}. 
Tree-based methods naturally handle heterogeneous data types and can split on categorical features without requiring heavy embedding hence, they remain relatively stable, whereas neural baselines exhibit greater sensitivity to this alignment with operational data. Deep learning on tabular data often struggles when the number of samples is moderate, when features vary in type and scale, and when the target function is irregular (non-smooth)~\cite{10.1016/j.inffus.2021.11.011}. 
This pattern suggests that semantic grounding modifies inductive biases rather than raw predictive accuracy.  
The magnitude of these effects also reveals a structural limitation of the underlying dataset: while \textsc{SALT-KG} introduces declarative semantics through OBKG, the relational scaffold inherited from \textsc{SALT} provides limited ontological depth (where the data exhibits only direct relationships between tables and fields, but does not capture higher-order abstractions such as class hierarchies, class-instance relationships, and rich expressivity through transitivity, inverse, reflexivity)  and cross-entity abstraction. 
As a result, available semantics cannot fully propagate through the relational topology, constraining the degree to which current models can internalize and exploit higher-order context. 
The modest gains observed indicate that semantics-aware learning also demands intrinsically structured datasets with a relational design that supports the emergence of semantic generalization.

\vspace{-5pt}
\section{Conclusion}
\vspace{-5pt}
By explicitly linking relational schema elements to declarative metadata in the OBKG, \textsc{SALT-KG} closes a critical gap between purely structural benchmarks for tabular learning and the semantics-grounded evaluation needed for the next generation of tabular foundation models. 
Empirical results, proffer a need for future work on architectures that unify relational, semantic, and linguistic understanding. The declarative semantics encoded in \textsc{SALT-KG} remain bounded by the descriptive depth of the underlying \textsc{SALT} dataset, where the
OBKG semantics are defined using existing enterprise schema and glossary, limiting the extent of cross-domain or hierarchical reasoning. 
\textsc{SALT-KG} thus provides the first benchmark for semantics-aware tabular learning and offers a foundation for systematically studying how contextual knowledge shapes predictive behavior. 
Future work should introduce ontological structure and broader cross-entity abstraction,  to enable models internalize higher-order context since semantics-aware learning depends not only on richer architectures but also on datasets whose relational design supports semantic generalization. Notably there is need to investigate the strength of the signals obtained from the OBKG and the effectiveness of the embedding representation used. We leave these as open research questions for
the community




\bibliography{bibliography}

\appendix
\section{Model Parameters}
\label{apdx:training-params}

\begin{table}[H]
\centering
\caption{Hyperparameters Used by Each Model}
\label{tab:hyperparameters}
\footnotesize 

\begin{minipage}[t]{0.48\textwidth}
\begin{tabular}{|lll|}
\hline
\textbf{Model} & \textbf{Parameter} & \textbf{Value} \\
\hline
\multirow{15}{*}{\textbf{AutoGluon}} 
& \multicolumn{2}{l|}{} \\
& \multicolumn{2}{l|}{\textit{NN\_TORCH Hyperparameters}} \\
& num\_epochs & 100 \\
& batch\_size & 8192 \\
& learning\_rate & 0.001 \\
& dropout\_prob & 0.1 \\
& num\_gpus & 1 \\
& & \\
& \multicolumn{2}{l|}{} \\
& \multicolumn{2}{l|}{\textit{FASTAI Hyperparameters}} \\
& epochs & 50 \\
& batch\_size & 4096 \\
& layers & {[}800, 400{]} \\
& num\_gpus & 1 \\
& & \\
& \multicolumn{2}{l|}{} \\
& \multicolumn{2}{l|}{\textit{General Training Parameters}} \\
& time\_limit & 7200s (2 hrs) \\
& holdout\_frac & 0.1 \\
& num\_bag\_folds & 3 \\
& presets & best\_quality \\
& \multicolumn{2}{l|}{} \\
& \multicolumn{2}{l|}{} \\
& & \\
& & \\
& & \\
& & \\
\hline
\end{tabular}
\end{minipage}
%
%
\begin{minipage}[t]{0.48\textwidth}
\begin{tabular}{|lll|}
\hline
\textbf{Model} & \textbf{Parameter} & \textbf{Value} \\
\hline
& & \\
\multirow{9}{*}{\textbf{GraphSAGE}} 
& epochs & 1000 \\
& patience & 5 \\
& num\_layers & 2 \\
& channels & 128 \\
& aggregation & sum \\
& normalization & batch\_norm \\
& learning\_rate & 0.0001 \\
& optimizer\_eps & 1e-8 \\
& tune\_metric & mrr \\
& & \\
\hline
& & \\
\multirow{11}{*}{\textbf{CARTE}} 
& num\_model & 1 \\
& random\_state & 42 \\
& n\_jobs & 1 \\
& loss & categorical\_ \\
&  & cross-entropy \\
& val\_size & 0.10 \\
& batch\_size & 4 \\
& early\_stopping &  \\
& \_patience & 5 \\
& learning\_rate & 0.0006 \\
& num\_layers & 3 \\
& dropout & 0.1 \\
& max\_epoch & 200 \\
& chunk\_size & 2048 \\
& & \\
\hline
\end{tabular}
\end{minipage}

\end{table}

\section{Dataset Statistics}
\label{apdx:data-stats}
This hybrid relational--graph structure supports new learning paradigms that jointly reason over transactional evidence and semantic context, combining tabular embeddings with graph-based reasoning. 
Table~\ref{tab:cds_summary} summarizes the composition and semantic augmentation introduced in SALT-KG.\\
\begin{table}[H]
\small
\centering
\caption{Summary of CDS Views infused in SALT-KG.}
\begin{tabular}{p{3.5cm}cc}
\toprule
\textbf{CDS View} & \textbf{Fields} & \textbf{Obj Node Types} \\
\midrule
\texttt{I\_SalesDocument} & 286 & 1 \\
\texttt{I\_SalesDocumentItem} & 412 & 1 \\
\texttt{I\_Customer} & 132 & 1 \\
\texttt{I\_Address\_2} & 94 & 1{,}954 \\
\texttt{I\_AddrOrgPostalAddress} & 66 & 1{,}954 \\
\bottomrule
\end{tabular}
\label{tab:cds_summary}
\end{table} 

\vspace{10pt}
\section{Sample KG Context: I\_SALESDOCUMENT}
\label{apdx:sample-context}
\begin{lstlisting}[style=json]
{
    "I_SALESDOCUMENT": {
        "name": "I_SALESDOCUMENT",
        "description": "Sales Document",
        "shortDescription": "This CDS view provides the prerequisites for answering questions about all relevant aspects of sales documents. Example business questions could include: What is the net value of a given sales document? Based on which document is a given sales order created? Which sales area does a given sales document belong to? Who is the sold-to party of a given sales document? When is a given sales order requested to be delivered? What is the overall processing status of a given sales document?",
        "details": "",
        "fields": [
            {
                "fieldName": "SALESDOCUMENT",
                "fieldDescription": "Sales Document",
                "fieldType": "abap.char",
                "fieldDetails": "",
                "dataElementDescription": "The number that uniquely identifies the sales document.",
                "targetColumn" : "SALESDOCUMENT"
            },
            {
                "fieldName": "SALESDOCUMENTTYPE",
                "fieldDescription": "Sales Document Type",
                "fieldType": "abap.char",
                "fieldDetails": "",
                "dataElementDescription": "A classification that distinguishes between different types of sales documents.",
                "targetColumn" : "SALESDOCUMENTTYPE"
            },
            {
                "fieldName": "SALESORGANIZATION",
                "fieldDescription": "Sales Organization",
                "fieldType": "abap.char",
                "fieldDetails": "",
                "dataElementDescription": "An organizational unit responsible for the sale of certain products or services.",
                "targetColumn" : "SALESORGANIZATION"
            },
            {
                "fieldName": "DISTRIBUTIONCHANNEL",
                "fieldDescription": "Distribution Channel",
                "fieldType": "abap.char",
                "fieldDetails": "",
                "dataElementDescription": "The way in which products or services reach the customer.",
                "targetColumn" : "DISTRIBUTIONCHANNEL"
            },
            {
                "fieldName": "ORGANIZATIONDIVISION",
                "fieldDescription": "Division",
                "fieldType": "abap.char",
                "fieldDetails": "",
                "dataElementDescription": "A way of grouping materials, products, or services.",
                "targetColumn" : "ORGANIZATIONDIVISION"
            },
            {
                "fieldName": "CREATIONDATE",
                "fieldDescription": "Record Creation Date",
                "fieldType": "abap.dats",
                "fieldDetails": "",
                "dataElementDescription": "",
                "targetColumn" : "CREATIONDATE"
            },
            {
                "fieldName": "CREATIONTIME",
                "fieldDescription": "Time at Which Record Was Created",
                "fieldType": "abap.tims",
                "fieldDetails": "",
                "dataElementDescription": "The time of day at which the document was posted and stored in the system.",
                "targetColumn" : "CREATIONTIME"
            },
            {
                "fieldName": "TRANSACTIONCURRENCY",
                "fieldDescription": "SD Document Currency",
                "fieldType": "abap.cuky",
                "fieldDetails": "",
                "dataElementDescription": "The currency that applies to the document.",
                "targetColumn" : "TRANSACTIONCURRENCY"
            },
            {
                "fieldName": "CUSTOMERPAYMENTTERMS",
                "fieldDescription": "Terms of Payment Key",
                "fieldType": "abap.char",
                "fieldDetails": "",
                "dataElementDescription": "Key for defining payment terms composed of cash discount percentages and payment periods.",
                "targetColumn" : "CUSTOMERPAYMENTTERMS"
            },
            {
                "fieldName": "SHIPPINGCONDITION",
                "fieldDescription": "Shipping Conditions",
                "fieldType": "abap.char",
                "fieldDetails": "",
                "dataElementDescription": "General shipping strategy for the delivery of goods from the vendor to the customer.",
                "targetColumn" : "SHIPPINGCONDITION"
            },
            {
                "fieldName": "INCOTERMSCLASSIFICATION",
                "fieldDescription": "Incoterms (Part 1)",
                "fieldType": "abap.char",
                "fieldDetails": "",
                "dataElementDescription": "Commonly used trading terms that comply with the standards established by the International Chamber of Commerce (ICC).",
                "targetColumn" : "INCOTERMSCLASSIFICATION"
            },
            {
                "fieldName": "BILLINGCOMPANYCODE",
                "fieldDescription": "Company Code to Be Billed",
                "fieldType": "abap.char",
                "fieldDetails": "",
                "dataElementDescription": "The company code represents an independent accounting unit.",
                "targetColumn" : "BILLINGCOMPANYCODE"
            },
            {
                "fieldName": "SALESGROUP",
                "fieldDescription": "Sales Group",
                "fieldType": "abap.char",
                "fieldDetails": "",
                "dataElementDescription": "A group of sales people who are responsible for processing sales of certain products or services.",
                "targetColumn" : "SALESGROUP"
            },
            {
                "fieldName": "SALESOFFICE",
                "fieldDescription": "Sales Office",
                "fieldType": "abap.char",
                "fieldDetails": "",
                "dataElementDescription": "A physical location that has responsibility for the sale of certain products or services within a given geographical area.",
                "targetColumn" : "SALESOFFICE"
            },
            "... [Status Fields - Collapsed for brevity]",
            "OVERALLSDPROCESSSTATUS, TOTALBLOCKSTATUS, OVERALLDELIVERYSTATUS",
            "OVERALLDELIVERYBLOCKSTATUS, OVERALLBILLINGBLOCKSTATUS",
            "OVERALLORDRELTDBILLGSTATUS, TOTALCREDITCHECKSTATUS",
            "CENTRALCREDITCHECKSTATUS, SALESDOCAPPROVALSTATUS",
            "... [Customer Fields - Collapsed for brevity]",
            "SOLDTOPARTY, CUSTOMERGROUP, CUSTOMERPRICEGROUP",
            "ADDITIONALCUSTOMERGROUP1-5, RETAILADDITIONALCUSTOMERGRP6-10",
            "CUSTOMERTAXCLASSIFICATION1-9",
            "... [Contract Fields - Collapsed for brevity]",
            "AGRMTVALDTYSTARTDATE, AGRMTVALDTYENDDATE",
            "SALESCONTRACTCANCLNREASON, SALESCONTRACTCANCLNPARTY",
            "SALESCONTRACTFOLLOWUPACTION, CONTRACTMANUALCOMPLETION",
            "... [Financial Fields - Collapsed for brevity]",
            "TOTALNETAMOUNT, CONTROLLINGAREA, COSTCENTER",
            "PRICEDETNEXCHANGERATE, ACCOUNTINGEXCHANGERATE",
            "SDPRICINGPROCEDURE, PRICINGDATE, PRICELISTTYPE",
            "... [Reference Fields - Collapsed for brevity]",
            "REFERENCESDDOCUMENT, REFERENCESDDOCUMENTCATEGORY",
            "PURCHASEORDERBYCUSTOMER, CUSTOMERPURCHASEORDERDATE",
            "EXTERNALDOCUMENTID, ACCOUNTINGDOCEXTERNALREFERENCE"
        ],
        "objNodeTypes": [
            {
                "objNodeTypeName": "SalesDocument",
                "objNodeTypeLabe": "Sales Document",
            }
        ]
    }
}
\end{lstlisting}


\end{document}